\documentclass[conference]{IEEEtran}
\IEEEoverridecommandlockouts
\usepackage{algorithm}
\usepackage{xcolor}
\usepackage{etoolbox}

\newtoggle{editmode}

\togglefalse{editmode}  

\let\origtextcolor\textcolor
\let\origcolor\color

\renewcommand{\textcolor}[2]{%
  \iftoggle{editmode}{%
    \origtextcolor{#1}{#2}%
  }{%
    \ifstrequal{#1}{red}{%
      \origtextcolor{black}{#2}%
    }{%
      \origtextcolor{#1}{#2}%
    }%
  }%
}

\renewcommand{\color}[1]{%
  \iftoggle{editmode}{%
    \origcolor{#1}%
  }{%
    \ifstrequal{#1}{red}{%
      \origcolor{black}%
    }{%
      \origcolor{#1}%
    }%
  }%
}

\usepackage{algorithmic}
\usepackage{cite}
\usepackage{amsmath,amssymb,amsfonts}
\usepackage{algorithmic}
\usepackage{graphicx}
\usepackage{subcaption}
\usepackage{pifont}
\usepackage{textcomp}
\usepackage[normalem]{ulem}
\usepackage{booktabs}
\def\BibTeX{{\rm B\kern-.05em{\sc i\kern-.025em b}\kern-.08em
    T\kern-.1667em\lower.7ex\hbox{E}\kern-.125emX}}
\usepackage{array}

\begin{document}

\title{\textcolor{red}{A Transformer Model for Predicting Chemical Products from Generic SMARTS Templates with Data Augmentation}\\
}

\author{\IEEEauthorblockN{Derin Ozer\textsuperscript{1}, Sylvain Lamprier\textsuperscript{1}, Thomas Cauchy\textsuperscript{2}, Nicolas Gutowski\textsuperscript{1}, Benoit Da Mota\textsuperscript{1}}
\IEEEauthorblockA{ \textit{ \textsuperscript{1}Univ Angers, LERIA, SFR MATHSTIC, F-49000 Angers, France} \\
\textit{ \textsuperscript{2}Univ Angers, CNRS, MOLTECH-ANJOU, SFR MATRIX, F-49000 Angers, France}\\
\{derin.ozer, sylvain.lamprier, thomas.cauchy, nicolas.gutowski, benoit.damota\}@univ-angers.fr} 
}

\maketitle

\begin{abstract}
The accurate prediction of chemical reaction outcomes is a major challenge in computational chemistry. Current models rely heavily on either highly specific reaction templates or template-free methods, both of which present limitations. To address these, this work proposes the Broad Reaction Set (\textit{BRS}), a set featuring 20 generic reaction templates written in \textit{SMARTS}, a pattern-based notation designed to describe substructures and reactivity. \textcolor{red}{Additionally, we introduce \textit{ProPreT5}, a \textit{T5}-based model specifically adapted for chemistry and, to the best of our knowledge, the first language model capable of directly handling and applying \textit{SMARTS} reaction templates. To further improve generalization, we propose the first augmentation strategy for \textit{SMARTS}, which injects structural diversity at the pattern level. Trained on augmented templates, \textit{ProPreT5} demonstrates strong predictive performance and generalization to unseen reactions. Together, these contributions provide a novel and practical alternative to current methods, advancing the field of template-based reaction prediction.}
\end{abstract}


\section{Introduction}
The accurate prediction of chemical reaction outcomes is an important task in chemistry as it allows the construction of organic synthesis routes. Given a set of reactive molecules i.e., reactants, the goal is to determine the outcome, the product. This task is especially challenging, requiring a thorough understanding of chemical substances, compound classes, reactions, underlying reactivity patterns, and reaction conditions. Such an understanding is fundamental for various applications, including: Drug discovery \cite{b0}, material science \cite{material_discovery}, and green chemistry \cite{green_chem}.

Organic chemistry synthesis route planning has usually been an expert-driven and rule-based approach. However, the advancement of machine learning has transformed the field of cheminformatics. Among the most promising techniques in this field are Transformer models \cite{b1}, originally developed for sequence-to-sequence tasks in Natural Language Processing (\textit{NLP}) \cite{b2}. The Transformer architecture has revolutionized \textit{NLP} by enabling models to understand context through self-attention mechanisms. This capability helps the models to learn to dynamically assign weights to different parts of the input data to produce more accurate and contextually relevant outputs. The Transformer architecture has also been successfully applied to the field of chemistry in various tasks, such as single-step chemical reaction prediction \cite{b3, augtrans, chemformer}, retrosynthesis \cite{chemformer, retro_1}, molecule generation \cite{b5, taiga, llamol}, and molecular property prediction \cite{bert_prop_pred, b6, molformer}. However, these models are highly dependent on the quality and diversity of the data used for training.

Molecular data can be represented in the form of strings, which are derived from a graph traversal of their molecular structure. This compact representation, known as the Simplified Molecular Input Line Entry System (\textit{SMILES}) \cite{b9.5} encodes molecules where each character represents an atom or bond, providing a linear representation useful for computational purposes and database storage. When the graph traversal follows the specific order established by the notation, the resulting \textit{SMILES} is canonical, and a molecule has only one canonical \textit{SMILES} but can have multiple non-canonical SMILES representations.

The SMILES notation can also be used to represent chemical reactions, encoding reactants (starting materials), reagents (substances that assist the reaction but do not transform), and products (final molecules produced). The \textit{SMARTS} \cite{smarts} notation, on the other hand, extends \textit{SMILES} by representing patterns of atoms and bonds, or substructures within a molecule. It allows the identification of functional groups in reactants, creating reaction templates that capture general reactivity patterns and improve the prediction and recognition of chemical behavior. 

The task of predicting reaction products studied in this paper is typically addressed using two different approaches: template-based methods, which rely on predefined rules (such as reactions represented by reaction \textit{SMILES} or \textit{SMARTS}), and template-free methods, which learn reactivity patterns directly from large datasets without relying on predefined rules or patterns.

Publicly available reaction datasets are obtained through data extraction from patents published by the United States Patent and Trademark Office (\textit{USPTO}) \cite{b9}. These reactions are represented using \textit{SMILES} representation along with their reactants and reagents. Previous work has created a sub-dataset called \textit{USPTO MIT} \cite{b10} by filtering it to 470,000 examples and splitting it into train/test/validation sets. \textit{USPTO MIT} datasets are widely used in the literature.

Since \textit{USPTO} datasets are derived from patents, they provide highly specific reactions tailored exclusively to particular reactant-product combinations. Furthermore, the chemical space explored using these datasets is limited to the information contained within the patented reactions. Consequently, we assume that relying solely on patents introduces bias during training, leaving a substantial portion of the chemical space unaccounted for. While previous work \cite{chemformer} has demonstrated great accuracy in predicting reaction outcomes using this dataset, the lack of real-world applications stemming from such models highlights a key limitation: models trained solely on the patented chemical space suffer from generalization issues and are unsuitable for practical applications \cite{uspto_arg1}. 

Such limitations and observations have led us to propose a new dataset called Broad Reaction Set (\textit{BRS}) that bridges the gap between regular expressions and overly specific reaction \textit{SMILES}. These reactions are expressed using the \textit{SMARTS} syntax and represent broader transformation patterns rather than single, specific reactions. This approach enables us to explore a more comprehensive portion of the chemical space, which is a fundamental step for the tasks of organic synthesis and molecule discovery. 

Reaction template datasets are often argued to be difficult to maintain due to the endless and unmanageable number of reactant-product combinations \cite{augtrans}, leading many to favor template-free methods as the future of the field. Maintaining templates for highly specific reactions, such as those in \textit{USPTO}, is undeniably challenging. In contrast, we argue that using generic reactions, as in the proposed \textit{BRS}, preserves the benefits of reaction templates while avoiding the associated maintenance issues.

To explore the potential of the proposed \textit{BRS}, we introduce \textit{ProPreT5}, a \textit{T5}-based model tailored for the reaction product prediction task. \textit{ProPreT5} is trained in both template-based and template-free settings using the standard \textit{USPTO MIT} dataset as well as a newly constructed dataset based on the generic \textit{BRS} reaction templates. The aim of this work is not to improve performance on \textit{USPTO MIT}, but to propose a more flexible and realistic alternative. For this reason, we adopt an economical training setup and use \textit{USPTO MIT} primarily as a sanity check. \textcolor{red}{In contrast, our main focus lies in training and evaluating the model on the more challenging dataset derived from \textit{BRS}. To support learning in this setting, we also propose a novel augmentation strategy for \textit{SMARTS} templates, designed to enhance generalization from a limited number of generic reactions.}

The contributions presented in this article are threefold:
\begin{itemize}
    \item We address the limitations of publicly available reaction datasets by introducing a novel generic reaction set \textit{BRS}, allowing for a broader exploration of the chemical space.
    \item We release \textit{ProPreT5}, an improved and highly flexible \textit{T5}-based model capable of generating accurate reaction products. \textcolor{red}{To the best of our knowledge, it is also the first large language model capable of handling and applying \textit{SMARTS} templates.}
    \item \textcolor{red}{We propose the first augmentation strategy for \textit{SMARTS} templates, using specialization, generalization, combinations, and permutations of atom patterns to introduce chemically consistent diversity and improve model robustness.}
\end{itemize}

\begin{table*}[!h]
\caption{Generic Reaction Patterns from the Broad Reaction Set (BRS)} 
\label{table1}
\begin{center}
\resizebox{1\textwidth}{!}{
\begin{tabular}{|c|>{\raggedright\arraybackslash}m{0.47\textwidth}|c|>{\raggedright\arraybackslash}m{0.45\textwidth}|}
\hline
\multicolumn{4}{|c|}{\textbf{SMARTS Notation}}\\
\hline
\textbf{\textit{\#}} & \textbf{Constructive Reactions} & \textbf{\textit{\#}} & \textbf{Destructive Reactions} \\
\hline
1 & \texttt{[\#6,\#7,\#8;h:1].[O,N,F,C:2]>>[\#6,\#7,\#8:1][O,N,F,C:2]} & 11 & \texttt{[\#6,\#7,\#8:1][O,N,F,C:2]>>[\#6,\#7,\#8;h:1]} \\
\hline
2 & \texttt{[O,N,C;h:1][O,N,C;h:2]>>[O,N,C:1]=[O,N,C:2]} & 12 & \texttt{[O,N,C:1]=[O,N,C:2]>>[O,N,C;h:1][O,N,C;h:2]} \\
\hline
3 & \texttt{[N,C;h2:1][N,C;h2:2]>>[N,C:1]\#[N,C:2]} & 13 & \texttt{[N,C:1]\#[N,C:2]>>[N,C;h2:1][N,C;h2:2]} \\
\hline
4 & \texttt{[C;h:1]=[N,C;h:2]>>[C:1]\#[N,C:2]} & 14 & \texttt{[C:1]\#[N,C:2]>>[C;h:1]=[N,C;h:2]} \\
\hline
5 & \texttt{[\#6,\#7,\#8;h:1]\ensuremath{\sim}[*:2]\ensuremath{\sim}[\#6,\#7,\#8;h:3]>> [\#6,\#7,\#8:1]1[*:2]\ensuremath{\sim}[\#6,\#7,\#8:3]1} & 15 & \texttt{[\#6,\#7,\#8:1]1[*:2]\ensuremath{\sim}[\#6,\#7,\#8:3]1>> [\#6,\#7,\#8;h:1]\ensuremath{\sim}[*:2]\ensuremath{\sim}[\#6,\#7,\#8;h:3]} \\
\hline
6 & \texttt{[\#6,\#7,\#8;h:1]\ensuremath{\sim}[*:2]\ensuremath{\sim}[*:4]\ensuremath{\sim}[\#6,\#7,\#8;h:3]>> [\#6,\#7,\#8:1]1[*:2]\ensuremath{\sim}[*:4]\ensuremath{\sim}[\#6,\#7,\#8:3]1} & 16 & \texttt{[\#6,\#7,\#8:1]1[*:2]\ensuremath{\sim}[*:4]\ensuremath{\sim}[\#6,\#7,\#8:3]1>> [\#6,\#7,\#8;h:1]\ensuremath{\sim}[*:2]\ensuremath{\sim}[*:4]\ensuremath{\sim}[\#6,\#7,\#8;h:3]} \\
\hline
7 & \texttt{[\#6,\#7,\#8;h:1]\ensuremath{\sim}[*:2]\ensuremath{\sim}[*:4]\ensuremath{\sim}[*:5]\ensuremath{\sim}[\#6,\#7,\#8;h:3]>> [O,N,C:1]1[*:2]\ensuremath{\sim}[*:4]\ensuremath{\sim}[*:5]\ensuremath{\sim}[\#6,\#7,\#8:3]1} & 17 & \texttt{[O,N,C:1]1[*:2]\ensuremath{\sim}[*:4]\ensuremath{\sim}[*:5]\ensuremath{\sim}[\#6,\#7,\#8:3]1>> [\#6,\#7,\#8;h:1]\ensuremath{\sim}[*:2]\ensuremath{\sim}[*:4]\ensuremath{\sim}[*:5]\ensuremath{\sim}[\#6,\#7,\#8;h:3]} \\
\hline
8 & \texttt{[\#6,\#7,\#8;h:1]\ensuremath{\sim}[*:2]\ensuremath{\sim}[*:4]\ensuremath{\sim}[*:5]\ensuremath{\sim}[*:6]\ensuremath{\sim} [\#6,\#7,\#8;h:3]>>[O,N,C:1]1[*:2]\ensuremath{\sim}[*:4]\ensuremath{\sim}[*:5]\ensuremath{\sim}[*:6] \ensuremath{\sim} [\#6,\#7,\#8:3]1} & 18 & \texttt{[O,N,C:1]1[*:2]\ensuremath{\sim}[*:4]\ensuremath{\sim}[*:5]\ensuremath{\sim}[*:6]\ensuremath{\sim}[\#6,\#7,\#8:3]1 >>[\#6,\#7,\#8;h:1]\ensuremath{\sim}[*:2]\ensuremath{\sim}[*:4]\ensuremath{\sim}[*:5]\ensuremath{\sim}[*:6]\ensuremath{\sim} [\#6,\#7,\#8;h:3]} \\
\hline
9 & \texttt{[\#6,\#7,\#8;h:1]\ensuremath{\sim}[*:2]\ensuremath{\sim}[*:4]\ensuremath{\sim}[*:5]\ensuremath{\sim}[*:6]\ensuremath{\sim}[*:7]\ensuremath{\sim} [\#6,\#7,\#8;h:3]>>[O,N,C:1]1[*:2]\ensuremath{\sim}[*:4]\ensuremath{\sim}[*:5]\ensuremath{\sim}[*:6] \ensuremath{\sim}[*:7]\ensuremath{\sim}[\#6,\#7,\#8:3]1} & 19 & \texttt{[O,N,C:1]1[*:2]\ensuremath{\sim}[*:4]\ensuremath{\sim}[*:5]\ensuremath{\sim}[*:6]\ensuremath{\sim}[*:7]\ensuremath{\sim} [\#6,\#7,\#8:3]1>>[\#6,\#7,\#8;h:1]\ensuremath{\sim}[*:2]\ensuremath{\sim}[*:4]\ensuremath{\sim}[*:5] \ensuremath{\sim}[*:6]\ensuremath{\sim}[*:7]\ensuremath{\sim}[\#6,\#7,\#8;h:3]} \\
\hline
10 & \texttt{[\#6,\#7,\#8;h:1]\ensuremath{\sim}[*:2]\ensuremath{\sim}[*:4]\ensuremath{\sim}[*:5]\ensuremath{\sim}[*:6]\ensuremath{\sim}[*:7]\ensuremath{\sim} [*:8]\ensuremath{\sim}[\#6,\#7,\#8;h:3]>>[O,N,C:1]1[*:2]\ensuremath{\sim}[*:4]\ensuremath{\sim}[*:5] \ensuremath{\sim}[*:6]\ensuremath{\sim}[*:7]\ensuremath{\sim}[*:8]\ensuremath{\sim}[\#6,\#7,\#8:3]1} & 20 & \texttt{[O,N,C:1]1[*:2]\ensuremath{\sim}[*:4]\ensuremath{\sim}[*:5]\ensuremath{\sim}[*:6]\ensuremath{\sim}[*:7]\ensuremath{\sim}[*:8] \ensuremath{\sim}[\#6,\#7,\#8:3]1>>[\#6,\#7,\#8;h:1]\ensuremath{\sim}[*:2]\ensuremath{\sim}[*:4]\ensuremath{\sim}[*:5] \ensuremath{\sim}[*:6]\ensuremath{\sim}[*:7]\ensuremath{\sim}[*:8]\ensuremath{\sim}[\#6,\#7,\#8;h:3]} \\
\hline
\end{tabular}
}
\end{center}
\vspace{1mm}
\footnotesize{\textit{Note:} The greater-than signs ($>>$)  separates reactants from products, while a dot (.) distinguishes individual molecules. $\#n$ represents any atom with the atomic number $n$, and specific letters indicate particular chemical environments; $:n$ is used to map and track specific subgraphs within the reaction. For more information on the SMARTS notation, refer to \cite{smarts}.}
\end{table*}

\section{Related Work}

Most of the recent literature on template-based approaches in reaction prediction has focused on graph-based models. \cite{localretro, graph2}. However, to the best of our knowledge, the potential of Transformer models for template-based sequence prediction has yet to be explored. 


Template-free, sequence-based models have achieved impressive results on the \textit{USPTO MIT} benchmark, holding the current state-of-the-art in the single-step product prediction task \cite{chemformer}. However, these models face significant challenges in terms of applicability. The \textit{USPTO MIT} dataset, derived from patented reactions, captures only a fraction of the diversity and complexity found in real-world organic synthesis, leaving large regions of the chemical space unexplored. As a result, while template-free models perform well in generating reaction outcomes, they struggle to generalize to the discovery of novel molecules or the prediction of reaction outcomes that fall outside the dataset \cite{uspto_arg1}. 

Notable advancements in this domain include \textit{Molecular Transformer} \cite{b3}, which used a smaller version of the base Transformers architecture \cite{b1}. The model was trained on both forward and backward tasks. This was the first application of Transformers to the single-step reaction prediction task. \textit{Augmented Transformer} \cite{augtrans} used an extensive data augmentation on input \textit{SMILES} and demonstrated that doing so increased the model's generalization. \textit{Chemformer} \cite{chemformer}, on the other hand, took things even further by proposing a much larger \textit{BART} \cite{bart} model with special pretraining. The pretraining included masking and data augmentation, making the model more robust and increasing prediction accuracy. While these template-free Transformer models have demonstrated remarkable success in reaction prediction, the exploration of template-based sequence-to-sequence models for this task remains an underexplored area, leaving room for further development in this domain.

\textcolor{red}{The lack of real-world applications of existing reaction prediction models reveals a critical weakness: their inability to generalize beyond narrow, benchmark-specific patterns.} This lack highlights the need for a dataset with generic reactions, offering a more versatile training ground for reaction prediction models. This would allow models to explore the entire chemical space, ultimately advancing the field beyond the limitations of current benchmarks.

\section{Datasets}
\textcolor{red}{Three datasets are used in this work: the \textit{USPTO MIT} dataset \cite{b10}; \textit{BRS-Base}, a dataset constructed using the proposed \textit{BRS} generic reaction templates, one of the key contributions of this work; and \textit{BRS-Aug}, a \textit{SMARTS}-augmented version of the base training dataset, generated using the novel augmentation strategy proposed in this study, a second major contribution.} The \textit{USPTO MIT} dataset contains reactions, along with the corresponding reactants, reagents, and products extracted from patents, represented in \textit{SMILES} notation. The \textit{USPTO MIT} dataset includes highly specific reaction templates, valid only for the exact combination of reactants, reagents, and products. It is divided into two subsets: \textit{USPTO MIT Mixed}, which does not distinguish between reactants and reagents and is considered a slightly more challenging problem, and \textit{USPTO MIT Separated}, which separates reagents from reactants. 

In this study, we worked with the \textit{USPTO MIT Mixed} dataset with the standard train/validation/test split of approximately 409K, 30K, and 40K examples, respectively.

\subsection{Proposed Reaction Set: Broad Reaction Set}
In this study, we introduce 20 new generic reaction templates, as shown in Table \ref{table1}. Ten of these reactions are constructive, and the other ten are destructive, represented in \textit{SMARTS} notation. These reactions are inspired by those used in an evolutionary algorithm named \textit{EvoMol} \cite{evomol}, which demonstrated efficiency in exploring the chemical space \cite{bbo}. These reactions serve as foundational building blocks for constructing reaction prediction datasets similar to \textit{USPTO}. Since these reactions can be applied to a wide range of molecules, they offer significant flexibility for use with publicly available or commercial molecular datasets.

Destructive reactions, the reverse of constructive reactions, enable a return to an earlier stage in the chemical space, offering the possibility to explore alternative pathways. Constructive and destructive reactions are symmetrical.

Here’s what each reaction from Table \ref{table1} does:
\begin{itemize}
    \item Reactions 1 \& 11: The constructive reaction (\#1) involves a molecule with an atom such as C, N, or O, bonded to hydrogen, reacting with a functional group containing O, N, F, or C. The functional group is added to the reactant. In contrast, the destructive reaction (\#11) removes a functional group containing O, N, F, or C and replaces it with a hydrogen atom.
    \item Reactions 2 \& 12: The constructive reaction (\#2) takes a molecule with a single bond between O, N, or C atoms which are bonded to H and forms a double bond instead. The destructive reaction (\#12) breaks a double bond between those same heavy atoms, replacing it with a single bond.
    \item Reactions 3 \& 13: The constructive reaction (\#3) takes a molecule with a single bond between N and C atoms, each bonded to two hydrogen atoms, and replaces this bond with a triple bond between these atoms, breaking the bond of one hydrogen atom for each. The destructive reaction (\#13) breaks the triple bond and adds a hydrogen atom to each heavy atom that forms the bond.
    \item Reactions 4 \& 14: The constructive reaction (\#4) takes a molecule with a C atom bonded to a H. This C atom contains a double bond with a N or C atom, which in turn is also bonded to at least one H. The reaction transforms the double bond into a triple bond. The destructive reaction (\#14) takes a molecule with a triple bond between a C atom and a N or C atom and breaks this triple bond into a double bond.
\end{itemize}

The remaining reactions focus on the creation or destruction of cycles of various sizes and will be explained together:
\begin{itemize}
    \item Reactions 5 -- 10~~\&~~15 -- 20: The constructive reactions (\#5 -- \#10) involve a molecule with a linear structure where an atom C, N or O bonded to H is connected to one (for reaction 5) up to six (for reaction 10) wildcard atoms (any atom), forming a cyclic structure by bonding the atom at position :1 with the atom at position :3 to close the ring. The destructive reactions (\#15 -- \#20) involve a molecule with a cycle of size 3 (for reaction 15) up to 8 (for reaction 20) and they break the cycle.
\end{itemize}

With their highly generic patterns, these reactions can be applied to a wide range of molecules, as well as different substructures within the same molecule. This makes the reaction set extremely flexible and well-suited for exploring chemical space. It is argued that these 20 reactions provide a solid foundation for starting with the simplest molecules and exploring a significant portion of the chemical space.

For simplicity, the transformations defined in these reactions are currently limited to atoms C, N, O, and F. However, this does not imply that the reactions cannot be applied to molecules containing other atoms. It simply means that only the reactivity of these four atoms is considered, and transformations will occur exclusively between them within the molecule. Additionally, reactions constructing cycles up to size 8 were defined. This limitation is partly due to challenges in defining reactions that can generate cycles of arbitrary size using SMARTS notation, but also because cycles larger than size 8 are rare. Moreover, chemists have proposed metrics that try to assess the synthesizability of a molecular graph, such as SAScore \cite{sascore}, where large cycles are penalized.

Despite these limitations, the defined reactions allow us to explore a significant portion of the chemical space. With minor modifications, these reactions can be extended to include additional atoms and cover a larger part of the chemical space, exceeding the limits set by the current definition. For the time being, the limitations we have set still enable us to explore the chemical space.

\subsection{Dataset Construction} \label{subsec_b}
\textcolor{red}{To construct the dataset used in this study, reactants were randomly sampled from two publicly available sources: \textit{EVO10} \cite{evo10}, which enumerates all possible molecules with up to 10 atoms of C, N, O, F, or S; and \textit{ChEMBL34} \cite{chembl}, a large-scale dataset of bioactive, drug-like molecules. \textit{ChEMBL34} was filtered to retain only molecules containing the same atom types as \textit{EVO10}, ensuring chemical consistency and relevance across sources.
The dataset was created using molecules sampled from these two sources. For each reaction in the template set, a molecule was randomly selected and evaluated for compatibility with the input pattern. If the structure matched, the reaction was executed using \textit{RDKit} \cite{rdkit}, an open-source cheminformatics library. If the reaction produced valid products, a series of filtering steps was applied to eliminate unrealistic outcomes commonly introduced by generic \textit{BRS} templates.
First, only canonical \textit{SMILES} were retained. Then, we applied the filtering criteria from \cite{evo10}: one filter removed molecules containing substructures not found in real-world compound datasets such as \textit{ChEMBL} and \textit{ZINC} \cite{zinc20}; the other excluded products containing Generic Cyclic Features (\textit{GCF}), molecular ring scaffolds not observed in those datasets. Products that passed these filters were deemed realistic. To promote structural diversity, we retained multiple products for the same reactant–reaction combination, allowing the model to learn that a single input may correspond to multiple plausible outcomes. This resulted in the generation of the \textit{BRS-Base} dataset consisting of 220K training, 10K validation, and 10K test examples. Much larger datasets can also be generated by applying the \textit{BRS} templates to additional molecular sources.
}
\textcolor{red}{\subsection{Data Augmentation Strategy for SMARTS Templates} \label{dataug}
In this study, a new data augmentation strategy for \textit{SMARTS} templates is presented. To the best of our knowledge, this is the first attempt to augment this notation. Acquiring high-quality reaction templates is challenging, and in such cases, as in many other machine learning domains, data augmentation becomes a valuable tool. While previous work has successfully applied augmentation techniques to molecular \textit{SMILES} representations, leading to improved performance in template-free models \cite{augmentation, augtrans, chemformer}, no such strategy has been proposed for \textit{SMARTS}.
Inspired by these efforts, we introduce a novel augmentation framework tailored to \textit{SMARTS} templates. This is particularly important in our setting, where the number of base templates is limited to just 20. Without augmentation, the model may struggle to grasp the syntax and semantics of the SMARTS language from such a small and rigid set, potentially resulting in a model that merely memorizes 20 isolated input–output transformations. Our augmentation strategy addresses this limitation by introducing chemically sound diversity into the templates, promoting generalization across reactions, and encouraging the model to learn the underlying semantics of the transformations rather than surface-level patterns.
}

\textcolor{red}{
The transformations applied to the \textit{SMARTS} templates fall into four main categories:
\begin{itemize}
    \item \textbf{Specialization:} In \textit{SMARTS}, atoms can be represented by their atomic numbers using the \# symbol. For example, \#6 matches any carbon atom. We specialize these general representations by replacing them with specific forms such as C (a carbon atom outside of a ring) or c (a carbon atom within a ring). This transformation restricts the template to apply only to cyclic or non-cyclic structures, introducing meaningful variation.
    \item \textbf{Generalization:} Conversely, specific atom types like C or c can be generalized back to their atomic number form, such as \#6, allowing the template to match a broader range of reactants. While the actual chemical context may vary, this transformation preserves the previous reaction mechanism and increases the model’s exposure to syntactic diversity.
    \item \textbf{Permutation:} Atoms listed within \textit{SMARTS} brackets, e.g., [\#7, \#8, \#6; h:1], can be permuted without changing the chemical meaning. We apply permutations both within atom groups and between bracketed groups. For example, [\#6, \#7, \#8; h:2].[O, N, F, C:1] can be swapped to [O, N, F, C:1].[\#7, \#8, \#6; h:2]. These permutations increase syntactic diversity while preserving reactivity.
    \item \textbf{Combination:} We generate multiple combinations of atom classes within \textit{SMARTS} patterns. For instance, starting from [\#6, \#7, \#8; h:1], we can create variants such as [\#8; h:1], [\#6, \#7; h:1], and others. These combinations further expand the expressiveness of the templates.
\end{itemize}
}
\textcolor{red}{
There are important considerations when applying these transformations. Except for permutation, whenever a transformation is applied to the input (left-hand side) template, the same transformation must also be applied to the output (right-hand side) template. This is essential because, in chemical reactions, the atoms present in the input molecules persist in the output molecules, and maintaining consistency ensures chemical validity. In contrast, permutation only alters the order of atoms or groups within the \textit{SMARTS} pattern without changing their identity or role, so it does not require a corresponding change in the output template. Each augmented template must be validated using \textit{RDKit} \cite{rdkit} to ensure chemical correctness. 
Using these augmentation strategies, the size of the training dataset is doubled: the original entries are retained, and their augmented template versions are added. This results in a training set of 440K examples, referred to as \textit{BRS-Aug}.
}

\begin{figure*}[!t]
    \centering
    \includegraphics[width=0.9\textwidth]{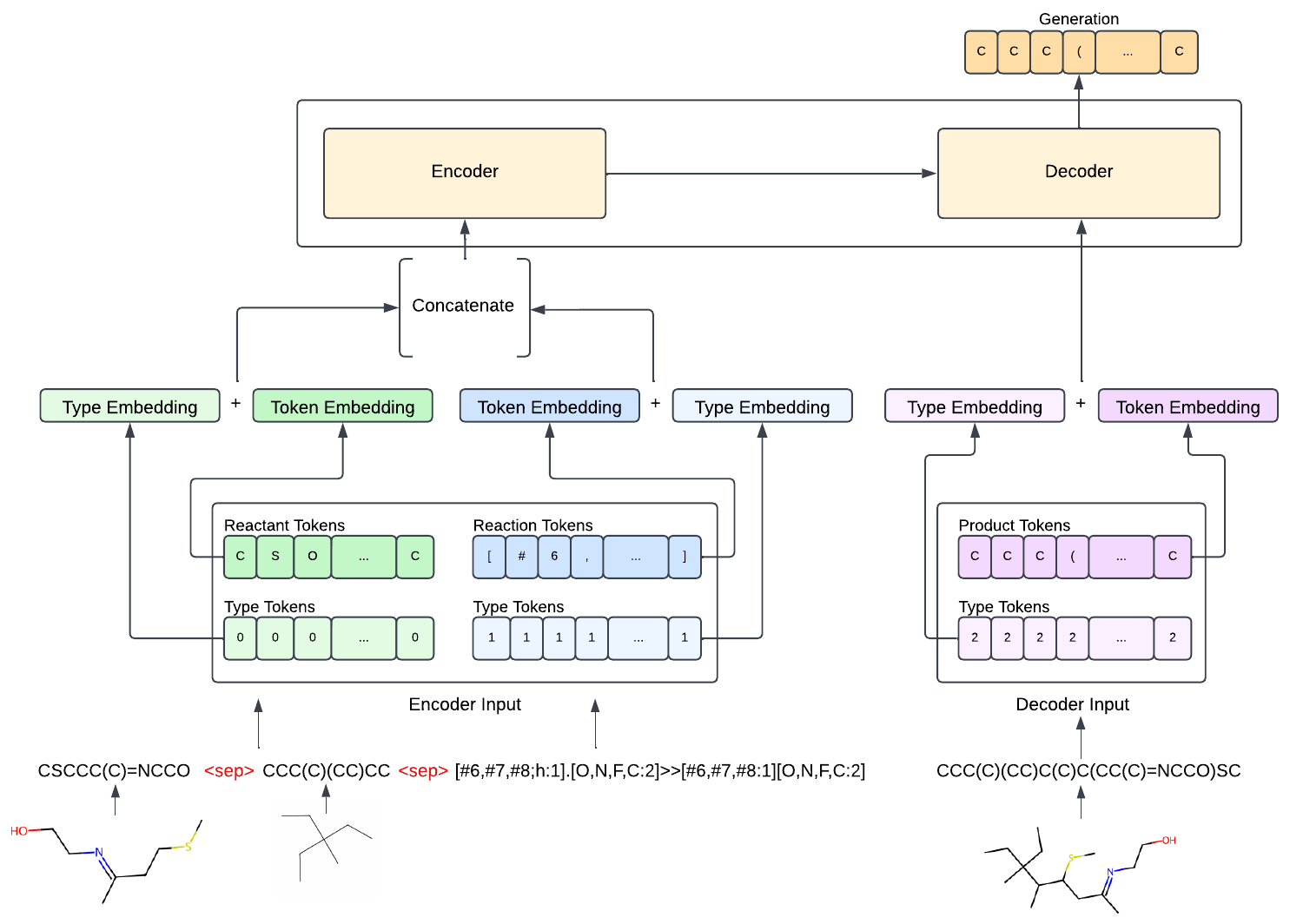}
    \caption{ProPreT5 Architecture.}
    \label{fig:ProPreT5}
\end{figure*}

\section{Method}
\subsection{Model and Implementation Details}
The general architecture of \textit{ProPreT5} is illustrated in Figure \ref{fig:ProPreT5}. A \textit{T5} model was chosen for its ease of use, relatively lightweight architecture, and ability to handle large datasets even with modest training resources. 
The model is very similar to the original \textit{T5} \cite{t5}, encompassing an encoder-decoder structure. The encoder and decoder blocks each consist of 6 layers with a hidden size of 512. Each block includes self-attention and feed-forward layers, with 8 attention heads. Feed-forward layers have an intermediate size of 2048 and use ReLU activation. In addition to these layers, the decoder includes cross-attention layers and masked self-attention. Relative positional embeddings are also used to capture token relationships.

Both the encoder and decoder share a vocabulary embedding layer with a vocabulary size of 243 tokens, as both the input and output use the same notation. The relatively small vocabulary size results from character-level tokenization, which enhances \textit{ProPreT5}'s flexibility. This approach allows most datasets to be used without retraining the model, unless a new character is added, unlikely given that the vocabulary covers a substantial portion of known chemistry. Character-level tokenization was selected to address the challenge of ensuring pretrained models have the correct tokens, as adding and fine-tuning new tokens is costly. 

As demonstrated in Figure \ref{fig:ProPreT5}, the reactants and the reaction are distinguished with a separation token used to separate both the reactants from one another and the reaction from the reactants. Additionally, type embeddings are applied to each input, following the approach in \cite{b5}. These embeddings distinguish between different input and output types, making it easier to incorporate new entry types, such as reagents or environmental conditions, without confusing the model. A separate trainable embedding layer is used to map these type tokens to embedding vectors. \textcolor{red}{\textit{ProPreT5} is trained separately to make predictions in both template-free and template-based settings: in the template-free setting, reaction templates are excluded from the encoder input; in the template-based setting, the model has access to the reaction template as part of the input.}

\subsection{Computational Resources and Training Setup}

The training and evaluation of \textit{ProPreT5} were completed on a high-performance computing cluster equipped with NVIDIA V100 GPUs with 32 GB of memory, providing sufficient capacity to handle large datasets. \textcolor{red}{\textit{ProPreT5} supports distributed training and was trained in parallel on 4 nodes, each equipped with 4 GPUs, for a total of 6 hours.} Unlike other models that require extensive computational resources and longer training times, this setup was relatively lightweight and time-efficient. Despite the short training duration, \textit{ProPreT5} achieved competitive performance.

\section{Results and Discussion} \label{results}
\textcolor{red}{\textit{ProPreT5} was trained in both template-free and template-based settings. While the \textit{USPTO MIT Mixed} benchmark is almost exclusively addressed in the template-free setting in the current literature, our use of this setting served primarily as a sanity check.} We applied template-free training on \textit{USPTO MIT Mixed} not to improve upon existing baselines, but to verify that \textit{ProPreT5} behaved as expected and achieved performance reasonably aligned with prior work. This step was essential to confirm that the model was functioning correctly before applying it to the more challenging and novel dataset, which is the primary focus of this study.

\subsection{Product Prediction on USPTO MIT}
\textcolor{red}{In the template-free setting using the \textit{USPTO MIT Mixed} dataset, we followed the standard approach in the literature where product prediction is performed without access to a reaction template. To improve generalization, we augmented the input molecules by a factor of four using non-canonical SMILES, as proposed in \cite{augmentation}; multiple non-canonical forms of each reactant were paired with the same product. This data augmentation is often employed in the existing literature \cite{b3, augtrans, chemformer}.} On the other hand, unlike some of the existing studies that rely on costly pretraining strategies, our model was trained from scratch, directly on \textit{USPTO MIT Mixed}.

Table \ref{uspto} presents the exact match accuracy of different models trained on the \textit{USPTO MIT Mixed} dataset. For this dataset, each reactant-reaction combination corresponds to exactly one product, so if the model did not generate that specific product, the prediction was considered incorrect.

As shown in Table \ref{uspto}, after only a few hours of training on a relatively economical setup, \textit{ProPreT5} achieved an exact match accuracy of 87.9\% in the template-free setting on the \textit{USPTO MIT Mixed} benchmark. This result is particularly notable given that the other three models required extensive training over hundreds of epochs. The accuracy was sufficient to confirm that the model was functioning as expected and ready to be applied to the more ambitious \textit{BRS} dataset.

\textcolor{red}{We also explored the template-based setting for the same benchmark. However, the other models could not be evaluated in this setting, as they lacked support for reaction template tokens. Introducing new tokens and retraining those models would have required significant effort and computational resources. However, \textit{ProPreT5} achieved a near-perfect accuracy of 99.8\% in the template-based setting on \textit{USPTO MIT Mixed}, supporting our earlier claim that the templates in this benchmark are highly specific and deterministic, effectively enumerating the atoms in the product.}


\begin{table}[h]
\caption{Prediction Accuracy on \textit{USPTO MIT Mixed}} 
\label{uspto}
\small
\color{red}
\begin{center}
\begin{tabular}{@{}lcc@{}}
\toprule
\textbf{Model} & \textbf{Template-based} & \textbf{Template-free} \\
\midrule
Molecule Transformer & - & 88.6\% \cite{b3} \\
Augmented Transformer & - & 90.0\% \cite{augtrans} \\
Chemformer & - & 90.9\% \cite{chemformer} \\
ProPreT5 & 99.8\% & 87.9\% \\
\bottomrule
\end{tabular}
\end{center}
\vspace{1mm}
\footnotesize{\textit{Note:} Template-based comparisons were omitted because none of the baseline models natively support reaction templates, and extending them with new tokens and retraining poses non-trivial challenges. Results with citations were directly taken from the corresponding papers.}
\end{table}

\subsection{Product Prediction on \textit{BRS}}
\textcolor{red}{Predicting reaction products for the dataset generated using the reactions from \textit{BRS} is inherently more complex than for the \textit{USPTO MIT Mixed} dataset. Unlike the highly specific templates in \textit{USPTO}, the \textit{BRS} templates are generic and do not explicitly enumerate the input or output molecules. Instead, they describe general transformation mechanisms that must be learned and generalized across diverse molecular contexts. The model must not only understand these transformations but also recognize that they can apply to a wide variety of molecules and to different regions within a single molecule. By design, this makes the task significantly more challenging.} Using \textit{BRS} templates, a single reactant-reaction combination can lead to multiple possible products, similar to how multiple correct translations can exist for a machine translation task. Therefore, in the case of \textit{BRS}, a generation is considered correct if the model predicted any one of the possible products. \textcolor{red}{It is important to note that, in this setting, no training sample shares the same reactant–reaction input as any test sample, a condition we carefully ensured.}

\textcolor{red}{The difficulty of the task increases further in the absence of reaction templates. Since a single molecule can serve as a reactant in many distinct reactions, the model lacks critical contextual cues and must rely on general transformation patterns. As a result, predictions become less reliable: the model may propose a chemically plausible product, but one that does not correspond to the specific reaction intended to produce the target product. This limitation is evident in the template-free results on the \textit{BRS-Base} dataset shown in Table \ref{table2}, where both models achieve prediction accuracies below 70\%.}

\textcolor{red}{For the template-free comparison, we selected \textit{Chemformer} \cite{chemformer}, as its code and trained weights were publicly available and it achieved the highest reported Top-1 accuracy on the \textit{USPTO MIT Mixed} benchmark. To evaluate its performance on our benchmark, we fine-tuned \textit{Chemformer} without reaction templates on the \textit{BRS-Base} training set. Using \textit{BRS-Aug} training set in the template-free setting is unnecessary, as the only difference from \textit{BRS-Base} lies in the reaction templates, which are excluded in this configuration. The underwhelming accuracy by both models in the template-free setting highlights a key limitation: in real-world scenarios where a single reactant can participate in a wide range of reactions, models trained without templates may lack the reliability needed for accurate prediction.}

\begin{table}[h]
\caption{Prediction Accuracy on \textit{BRS}}
\label{table2}
\small
\color{red}
\begin{center}
\begin{tabular}{@{}c|c|cc@{}}
\toprule
\textbf{Training} & \textbf{Reaction Template} & \textbf{Chemformer} \cite{chemformer} & \textbf{ProPreT5} \\
\midrule
BRS-Base & Template-free    & 69.2\%  & 68.8\% \\
BRS-Base & Template-based  & - & 91.2\% \\ 
BRS-Aug  & Template-based   & - & 95.1\% \\
\bottomrule
\end{tabular}
\end{center}
\vspace{1mm}
\footnotesize{\textit{Note:} Comparisons were not provided for the template-based versions, as \textit{Chemformer} \cite{chemformer} does not support reaction templates by default, and adding new tokens followed by retraining the model would be prohibitively expensive. The template-based result on \textit{BRS-Base} was obtained by fine-tuning \textit{Chemformer} on top of its existing training on the \textit{USPTO MIT Mixed} dataset. The reported accuracies are based on the \textit{BRS-Base} test set, while template augmentation (\textit{BRS-Aug}) is applied only during training to improve prediction performance.}
\end{table}

\textcolor{red}{When it comes to template-based prediction, which is the main challenge addressed in this paper, we aim to propose a language model capable of handling and understanding the \textit{SMARTS} notation. This capability is essential not only for accurate product prediction but also for enabling a variety of downstream tasks in the future. To evaluate this, we first trained the model on the non-augmented version of the dataset \textit{BRS-Base}. The model achieved a solid accuracy of 91.2\% on the test set. However, given the limited number of reaction templates, there is a risk that the model simply memorizes the reactions rather than learning the underlying transformation rules encoded in the \textit{SMARTS} notation.}

\textcolor{red}{To address this concern, we trained the model on the augmented version of the training set \textit{BRS-Aug}. This version includes multiple syntactic variants of each reaction template as explained in detail in Section \ref{dataug}, providing greater diversity and encouraging the model to generalize. As a result, the model's accuracy improved to 95.1\% on the same test set, indicating that the augmentation strategy indeed helps the model capture a more nuanced understanding of the reaction rules.}

\textcolor{red}{This effect becomes particularly evident when we assess the model's robustness by removing a specific reaction template from the training set, specifically, Reaction 5 from Table \ref{table1}, along with all its augmented variants. As shown in Table \ref{tablecomp}, the model trained on the non-augmented \textit{BRS-Base} dataset without the Reaction 5 suffers a significant drop in performance when evaluated on that same reaction, suggesting that it had overfitted to the limited examples without truly learning the \textit{SMARTS} syntax. In contrast, the model trained with augmented templates maintains much better performance, providing strong evidence that the augmentation strategy enhances generalization and improves the model’s ability to learn the \textit{SMARTS} language in a meaningful way.}

\begin{table}[h]
\caption{Prediction Accuracy of \textit{ProPreT5} on Reaction 5 Removed from Training Set}
\label{tablecomp}
\small
\color{red}
\begin{center}
\begin{tabular}{@{}c|c@{}}
\toprule
\textbf{Training Subset}  & Accuracy on Reaction 5 \\
\midrule
BRS-Base without Reaction 5  & 23.3\% \\ 
BRS-Aug without Reaction 5  & 54.7\% \\
\bottomrule
\end{tabular}
\end{center}
\vspace{1mm}
\footnotesize{\textit{Note:} Reaction 5 and all its augmented variants were removed from the training sets. The table reports the model's accuracy on test samples involving Reaction 5.}
\end{table}

\section{Conclusion}
In this study, we introduced the Broad Reaction Set (\textit{BRS}), a novel collection of generic reaction templates designed to provide a more realistic and versatile benchmark for reaction prediction. This contribution addresses key limitations of widely used datasets such as \textit{USPTO}, which rely on overly specific reaction templates and therefore limit real-world applicability, an essential requirement in cheminformatics.

\textcolor{red}{We also proposed a novel augmentation strategy for the \textit{SMARTS} notation, which, to the best of our knowledge, has not been explored previously. This strategy was applied to the \textit{BRS} templates, resulting in two training sets: \textit{BRS-Base}, using the original templates, and \textit{BRS-Aug}, incorporating augmented variants.} While developed for the 20 generic reactions presented in this study, the augmentation strategy is generalizable and can be applied to other \textit{SMARTS}-based reaction templates.

\textcolor{red}{To evaluate the impact of the proposed datasets, we introduced \textit{ProPreT5}, a flexible \textit{T5}-based model for reaction product prediction. Its flexibility stems not only from being trained on generic reactions but also from its character-level tokenization, which enables seamless adaptation to other datasets without requiring costly retraining. \textit{ProPreT5} is a major contribution, as it is the first language model capable of directly handling and applying the \textit{SMARTS} notation. This opens the door to new applications: thanks to its strong predictive performance, \textit{ProPreT5} can serve as a faster and more scalable alternative to \textit{RDKit} for product prediction, particularly in scenarios where batching is required, something that \textit{RDKit}'s template application does not support efficiently. Although its accuracy is not yet perfect, \textit{ProPreT5} offers a compelling balance of speed and flexibility. Furthermore, it provides a strong baseline for future downstream tasks involving \textit{SMARTS}-based reasoning.}

It is important to emphasize the broader utility of developing models capable of interpreting chemical reactions from generic templates. By learning the underlying rules of chemistry, such models are better prepared for downstream tasks that require a deeper understanding of reaction mechanisms. Moreover, unlike rule-based tools such as RDKit, which often fail when faced with noisy or imperfect input data, these models offer greater flexibility and generalizability. Their ability to handle variability and incomplete information makes them valuable assets for real-world applications where input quality cannot always be guaranteed.

Finally, we showed that satisfactory results can be obtained with lightweight training configurations, avoiding the need for extensive pretraining. In future work, we plan to extend this approach to multi-step synthesis planning, leveraging the template-based framework of \textit{ProPreT5}. We also aim to further improve prediction accuracy in order to minimize error propagation in multi-step workflows.

\section*{Acknowledgment}
This work was supported by the University of Angers, and the French Ministry of Education and Research (JL PhD grant). This work was performed using HPC resources from GENCI-IDRIS (Grant 2024-AD011014840R1)

\section*{Code and Data Availability}
The code and the data for this work will be made available in the future as they cannot be shared at this time to maintain the anonymity of the submission.

\end{document}